\documentclass[aps,twocolumn,pre,showpacs,superscriptaddress,groupedaddress,nofootinbib]{revtex4-2}  
\usepackage{amssymb}   
\usepackage{amsmath}
\usepackage{multirow}
\usepackage{makecell}
\usepackage{graphicx}  
\usepackage{xcolor}
\usepackage{soul}
\usepackage{hyperref}

\begin{document}

\title{Optimal Machine Intelligence at the Edge of Chaos}

\author{Ling Feng}\thanks{}
\affiliation{Institute of High Performance Computing, A*STAR, 138632 Singapore}
\affiliation{Department of Physics, National University of Singapore, 117551 Singapore}
\author{Lin Zhang}
\affiliation{Department of Physics, National University of Singapore, 117551 Singapore}
\author{Choy Heng Lai}
\affiliation{Department of Physics, National University of Singapore, 117551 Singapore}
\affiliation{Centre for Quantum Technologies, National University of Singapore, 117543 Singapore}
\affiliation{NUS Graduate School for Integrative Sciences and Engineering, 119077 Singapore}

	\date{
	 \today \\
}

\begin{abstract}
It has long been suggested that the biological brain operates at some critical point between two different phases, possibly  order and chaos. Despite many indirect empirical evidence from the brain and analytical indication on simple neural networks, the foundation of this hypothesis on generic non-linear systems remains unclear. Here we develop a general theory that reveals the exact edge of chaos is the boundary between the chaotic phase and the (pseudo)periodic phase arising from Neimark-Sacker bifurcation. This edge is analytically determined by the asymptotic Jacobian norm values of the non-linear operator and influenced by the dimensionality of the system. 
The optimality at the edge of chaos is associated with the highest information transfer between input and output at this point similar to that of the logistic map. As empirical validations, our experiments on the various deep learning models in computer vision demonstrate the optimality of the models near the edge of chaos, and we observe that the state-of-art training algorithms push the models towards such edge as they become more accurate. We further establishes the theoretical understanding of deep learning model generalization through asymptotic stability.
\end{abstract}

\maketitle


\section{Introduction}

There has been abundant suggestive evidence that many natural systems operate around the critical point between order and disorder \cite{munoz2018colloquium}. In particular the brain activities exhibit various spatiotemporal patterns of scale-invariance, which resemble that of critical phase transitions in statistical mechanics~\cite{Beggs11167, Fraiman2009}. However, direct evidence of criticality on these systems are limited due to the practical limitations to measure their exact microscopic dynamics to validate the claim. On the theoretical front, self-organized criticality~\cite{bak59self} was initially proposed to explain the prevalence of scale-invariance in nature. Certain complexity measures are shown to be maximized at the edge of chaos for dynamical systems through simulations~\cite{Crutchfield2008}. 
Simple computer models have been able to demonstrate that certain non-linear systems at criticality possess maximal adaptivity and information processing capability, leading to the hypothesis that living systems optimize themselves towards the critical state to maximize adaptivity and survival~\cite{HUBERMAN1986376}. Another notion of criticality called critical branching in the brain observed in the brain~\cite{beggs2003neuronal} was studied using computer simulations, showing that maximal number of metastable states are present at this critical phase, and this was proposed as the explanation for the critical brain observation~\cite{haldeman2005critical}. But to date, a clear understanding for this edge of chaos hypothesis is yet to be established. 

Extending this natural phenomenon to  artificial neural networks,   correlation between edge of chaos and computational power has been observed~\cite{bertschinger2004real, LEGENSTEIN2007323}. Studies using spin glasses models~\cite{sherrington1975solvable} have made numerous contributions in understanding the information processing power of single layer fully connected neural networks~\cite{nishimori2001statistical}, mostly independent from the order-chaos framework. The first theoretical result on the transition to chaos in a single layer neural network with random weights was established in~\cite{sompolinsky1988chaos}, and was later extended to show that such system has optimal sequence memory near the edge of chaos~\cite{Toyoizumi:2011aa}. With the recent development in deep neural networks that can solve real world problems, the edge of chaos hypothesis can now be rigorously tested on machine intelligence. 

Here we first establish a theory that applies to generic non-linear systems including complex deep neural networks, proving their maximal information processing power at the edge of chaos that can be analytically demonstrated. The theoretical results are further validated using various deep learning models in computer vision on benchmark datasets.

\section{Theory}

We start with establishing the theory for generic dynamical operators, and then extending it to deep neural networks. For any discrete dynamical operator of the generic form: $\pmb x_{t+1}=\pmb f(\pmb x_t)$, where $\pmb x_t$ is a vector of $N$ dimensions, at long enough time of $t \to \infty$ the systems will evolve to its asymptotic attractor(s) denoted as $\pmb x^* = \pmb f(\pmb x^*)$. In general the attractor can be deterministic or chaotic. Hence a generic expression for the attractor is:
\begin{align}
\pmb x^* = \pmb \mu + \pmb\xi,
\end{align}
where $\pmb \mu$ is the deterministic component, and $\pmb \xi$ is the remaining chaotic component described by a random variable with mean $\pmb 0$ and standard deviation $\pmb \sigma$. If the attractor $\pmb x^*$ is deterministic, it simply means the chaotic component  $\pmb \xi = \pmb 0$, i.e. $\pmb \sigma = \pmb 0$. Hence the transition from deterministic to chaos happens when $\pmb \sigma$ starts to deviate from $\pmb 0$ -- a second-order phase transition with order parameter $\pmb \sigma$. For high dimensional systems where $N$ is large, employing Landau's theory we have the following result near the transition point (Sec.~\ref{Sec:general}):
\begin{align}
\|\pmb \sigma\|^2 \approx \frac{1}{N} \|\pmb J^*\|^2 \|\pmb \sigma\|^2,
\label{Eqn:self_sigma}
\end{align}
where $\pmb J^*$ is the Jacobian matrix of the operator $\pmb f$ evaluated at the asymptotic value $\pmb \mu$, $\|\pmb \sigma\|$ is the Euclidean norm, while $\|\pmb J^*\|$ is the Frobenius norm of $\pmb J^*$. We will denote the later as Jacobian norm for simplicity in the rest of the paper. Therefore the transition from deterministic/order phase to chaos is at the following critical threshold:
\begin{align}
\frac{1}{\sqrt{N}} \|\pmb J^*\| =1.
\label{critical}
\end{align}
If the operator $\pmb f$ is the special case of a single-layer fully connected neural network with random weights and $\tanh$ activation, one then recovers from Eqn.~\ref{critical} the known analytical result
for its order-to-chaos transition boundary~\cite{sompolinsky1988chaos} that is the same as the A-T line in spin glass~\cite{almeida1978stability}. Yet for many practical artificial neural networks, especially the modern architectures that are highly complex, analytical result can be extremely difficult to achieve. But this threshold can be numerically accessed with relative ease as we show later.

\begin{figure*}
  \centering
{\includegraphics[width=17cm]{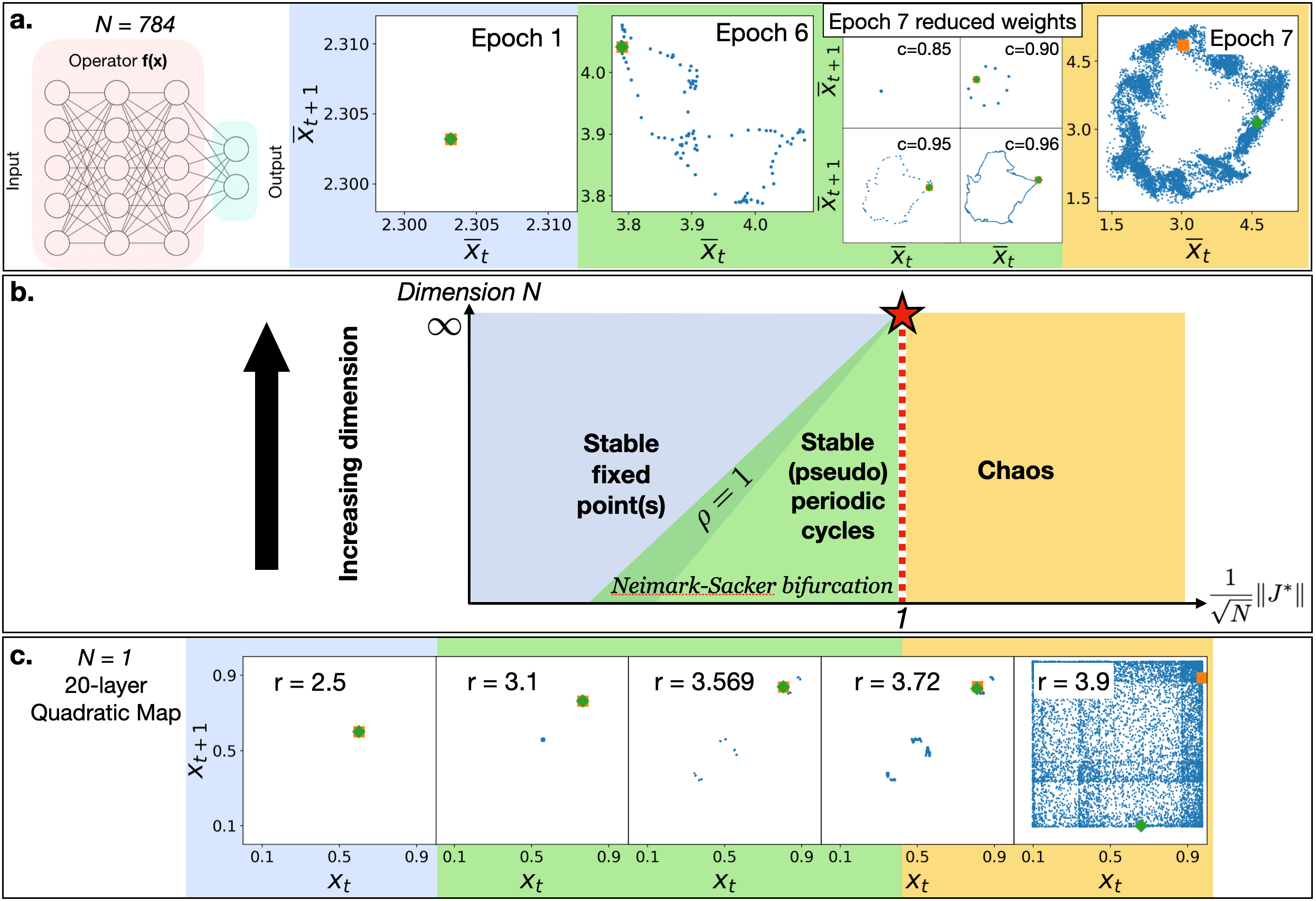}}
  \caption{{\bf Three phases of generic non-linear dynamical systems.} Blue background blocks refer to stable fixed point(s) phase, green for (pseudo)periodic phases, and orange for chaotic phases. 
  A. High dimension system. Poincare plot of the attractor of a multi-layer (2 hidden layer) fully connected neural network trained on the FashionMNIST dataset. The first 3 layers of the network is used as the dynamical operator $\pmb f$ as illustrated in the left diagram. Here $\overline{x}_t$ is the projection of $\pmb x_t$ onto an arbitrary fixed direction since $\pmb x_t$ is high dimensional. Epoch 6 and 7 have the best trained models with saturating test accuracy and lowest test loss throughout the whole training process,  and both have Jacobian norm close to the edge of chaos according to Eqn.~\ref{critical}. The 4 small figures in `Epoch 7 (reduced weights)' are the attractors from reducing every weight of the model in epoch 7 to a fraction of $c$. They exhibit a {\it Neimark-Sacker} bifurcation from single stable fixed point to periodic cycles with increasing cycle lengths.
    B. Phase diagram of generic non-linear dynamical systems. The red dotted line represents the edge of chaos in Eqn.~\ref{critical}, separating the (pseudo)periodic cycle phase and the chaotic phase. $\rho=1$ is the boundary separating the stable fixed point(s) phase and the (pseudo)periodic cycle phase, with the two phases overlapping since $\rho=1$ can only be described probabilistically.
  C. Low dimensional system. Dynamical operator $\pmb f$ is a 20 layer logistic map defined as: $x_{t+1}=f(x)=g^{20}(x_t)$, where $g(x)=rx(1-x)$. The Poincare plots of the attractors show the well-known period doubling bifurcation process before reaching the chaotic phase.
  In both A and C, the orange square and the green diamond represent the final values of 2 initially nearby points after $t=1000$ iterations. If they overlap, it means the system is in order phase; otherwise the system is in the chaotic phase.}
\label{Fig:phase_diagram}
\end{figure*}

In the thermodynamic limit of $N\to\infty$, the complex high non-linearity leads to high variance in the elements of the Jacobian matrix, while their mean value is minimal.
Hence, the critical threshold condition in Eqn.~\ref{critical} is  equivalent to $\pmb J^*$ being a random matrix whose elements are i.i.d. random variables with mean zero and variance $1/N$. Hence by random matrix theory, the eigenvalues of $\pmb J^*$ are distributed uniformly within a  circle of radius $1$ on the complex plane~\cite{tao2010random}. The largest absolute value of the eigenvalues, also known as the spectral radius $\rho$ consequently takes on the value $\rho=1$. 
In the ordered phase $\rho <1$ due to $\frac{1}{\sqrt{N}} \|\pmb J^*\| <1$, such that every fixed point $\pmb x^*$ has to be locally stable and no unstable fixed point exist, leading to a single stable fixed point in the asymptotic state. The loss of stability of this unique fixed point happens at the same time when it becomes a chaotic attractor at $\rho=\frac{1}{\sqrt{N}} \|\pmb J^*\| =1$, labeled by the red star in Fig.~\ref{Fig:phase_diagram}B.

However, for finite system dimension $N$, the two conditions $\rho=1$ and $\frac{1}{\sqrt{N}} \|\pmb J^*\|=1$ do not happen at the same time. In general $\rho$ is only related to $\frac{1}{\sqrt{N}} \|\pmb J^*\|$ probabilistically, with high probability that $\rho=1$ when $\frac{1}{N} \|\pmb J^*\|^2<1$ and $\rho>1$ when $\frac{1}{N} \|\pmb J^*\|^2=1$. We label the space between these two boundaries as the green region in Fig.~\ref{Fig:phase_diagram}B. 
In general, $\rho$ increases and becomes further away from $\frac{1}{\sqrt{N}} \|\pmb J^*\|$ as the dimension of the system dimension $N$ decreases~\cite{EDELMAN1997203}. This leads to a wider green region for low dimensions in Fig.~\ref{Fig:phase_diagram}B. At the boundary between the stable fixed point phase and the green phase, $\rho$ crosses the value 1, leading to a Neimark-Sacker bifurcation, such that the stable fixed point bifurcates into a (pseudo)periodic cycle. At the same time $\frac{1}{\sqrt{N}} \|\pmb J^*\|<1$ means such cycles are deterministic and stable. As $\frac{1}{\sqrt{N}} \|\pmb J^*\|$ increases towards 1, the spectral radius $\rho$ also increases(probabilistically) further away from 1, leading to larger periodic cycles, until chaos sets in at $\frac{1}{\sqrt{N}} \|\pmb J^*\|=1$. This process can be clearly observed for a high dimensional multi-layer neural network trained on real image dataset  (Fig.~\ref{Fig:phase_diagram}A).

Note that for very low dimensions, Eqn.~\ref{critical} is no longer accurate in capturing the transition point to chaos, because such transition is from the wide periodic cycle phase that can not be approximated as the stable fixed point phase. Hence, one needs to use the geometric mean of the asymptotic Jacobian norm $\frac{1}{\sqrt{N}} \overline{\|\pmb J^*\|}_c=1$ instead to determine the boundary between this periodic phase and  chaos. This quantity is essentially related to the finite time estimation of the maximal Lyapunov exponent $\gamma \approx \ln (\frac{1}{\sqrt{N}} \overline{\|\pmb J^*\|})$ that characterizes the system's stability (Sec.~\ref{Sec:lya}).
As an illustration, the well-known 1-dimensional  logistic map $f(x)=rx(1-x)$ exhibits the same process of increasing periodic cycle length through period doubling bifurcation in a wide range of  $r$ values~\cite{GREBOGI632,feigenbaum1976universality}, until it reaches  $\frac{1}{\sqrt{N}} \overline{\|\pmb J^*\|}=1$ at $r\approx 3.57$ (Fig.~\ref{Fig:phase_diagram}C). The spectral radius $\rho$ in this case is the absolute derivative at the non-trivial fixed point $x=\frac{r-1}{r}$, i.e. $\rho = |2-r|$, and the Lyapunov exponent is exactly  $\gamma = \ln\overline{J^*}$ where $\overline{J^*}$ is the geometric mean over the derivatives at all points in the attractor $x^*$ (Sec.~\ref{Sec:lya}).

As a generic feature of non-linear dynamical operators, the periodic cycles directly relate to the information processing power of the operator $\pmb f$. Assuming the asymptotic periodic cycle has length $L$, that means there can be up to $L$ different asymptotic outputs from the dynamics regardless of the initial point $\pmb x_0$, each with a different lag in the same cycle. In other words, there are $L$ metastable cyclic states of $\pmb f$, and they are the most numerous at the edge of chaos since the cycle length $L$ is the largest due to $\rho$ being the largest before chaos sets in. 
 Hence, the maximum amount of information that the dynamics  can generate at its asymptotic state is simply $\log_2 L$. Once it enters the chaotic phase, the $L$ different states are no longer distinctively resolved, leading to decrease in the information. This can be clearly illustrated through the input/output mutual information in the 20-layer logistic map model (Sec.~\ref{Sec:log_map}). In high dimensional systems, mutual information is hard to estimate reliably, but our theoretical result expect the same pattern.

The maximal information in the asymptotic states also indirectly infers the maximal information processing power with a single operation by $\pmb f$, measured in terms of the mutual information $I(x_0,\pmb f(x_0)) = I(x_0,x_1)$. Since the  information in the asymptotic states at the edge of chaos is $I(x_0, x_\infty)\gg 0$, it means the mutual information $I(x_0, x_1)$ is likely to be maximal when $\pmb f$ is at the edge of chaos, and this can be validated for the logistic map model shown in Fig.~\ref{Fig:Log_maps}. Note that such slow decay also indicates the memory lifetime of generic non-linear dynamical operators diverges to infinity near the edge of chaos. Such memory lifetime was studied for the special case of single layer fully connected neural network~\cite{Toyoizumi:2011aa} and under driven dynamics~\cite{Schuecker:2018aa}, and here we give the theoretical explanation for generic non-linear systems.

\section{Validation on deep neural networks}

Since the information processing power is the most optimal at the edge of chaos for a generic operator $\pmb f$, a neural network operator trained to find various patterns and distinguish different categories of data is supposed to be the most optimal near the edge of chaos. We carry out  experiments on the various computer vision models to confirm this hypothesis, by evaluating their stability phases vs. model performance during the training process. 
 The only constraint on the neural network operator  $\pmb f$  for stability analysis is that it needs to have the same dimension for its input and output, such that it can be used as a dynamical operator to evaluate $\pmb J^*$.  Therefore, we design every neural network such that a hidden layer close to the final layer has the same output dimension as the  images. The bulk of the neural network that does not include the final low dimensional output layer is used as dynamic operator $\pmb f$ for stability/chaos analysis. Such structural design does not have any impact on the accuracy of the various models (Sec.~\ref{Sec:empirical}).
To follow the best practice, we always use ReLU activation functions except for the final output layer, and train the networks using Adam optimizer~\cite{kingma2014adam}. 

\begin{figure*}
  \centering
{\includegraphics[width=11cm]{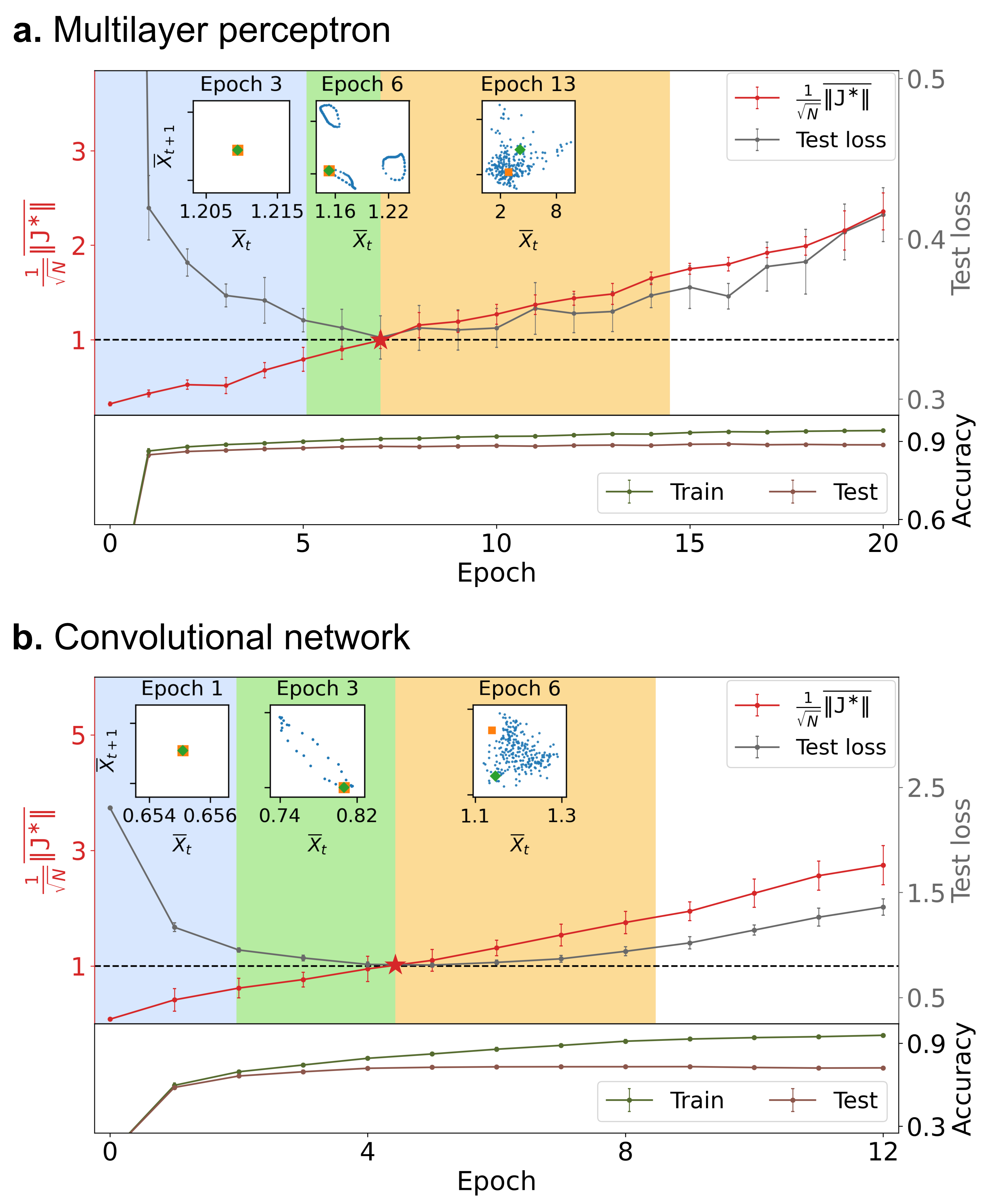}}
  \caption{{\bf Optimal neural network models near the edge of chaos.} A. The network consists of 2 hidden fully connected layers and dataset used is Fashion MNIST~\cite{xiao2017fashionmnist}. During the 20 epochs of training, the model transitions from the single fixed point phase to the periodic cycles and then the chaotic phase as labelled by the corresponding background colors. The insets are the Poincare plots similar to those in Fig.~1, illustrating the different phases. The model is the most optimal at epoch 7 (red star) as it has the lowest test loss and saturating test accuracy. The Jacobian norm $\frac{1}{\sqrt{N}} \overline{\|\pmb J^*\|}$ at this epoch is around 1, which corresponds to the edge of chaos. B. Convolutional network trained on CIFAR10 dataset~\cite{krizhevsky2009learning}. 
  Similar phase transition patterns are observed and the most optimal epoch is also found to be near the edge of chaos at epoch 5. Note that due to the unbounded activation ReLU used in both networks, there is an additional divergent phase with white background on the right of both subfigures. The error bars represent the standard deviation over 10 experiments. The details of the experiments are described in Sec.~\ref{Sec:empirical}.  
  }
\label{Fig:mlp_cnn}
\end{figure*}

At every epoch during the training process, we evaluate the normalized Jacobian norm $\frac{1}{\sqrt{N}} \overline{\|\pmb J^*\|}$ of the neural network operator at that epoch, and observe its relation with the test accuracy and loss. For the simple architecture of multi-layer perception that has multiple fully connected layer in sequence, the model is the most optimal with lowest test loss when $\frac{1}{\sqrt{N}} \overline{\|\pmb J^*\|}\approx 1$ as shown in Fig.~\ref{Fig:mlp_cnn}A and~\ref{Fig:appendix_mlp}, i.e. the edge of chaos. This validates our theoretical prediction based on the Jacobian norm. Extending this to the more complex convolutional neural networks, the model is also found to be the most optimal at the edge of chaos as shown in Fig.~\ref{Fig:mlp_cnn}B and \ref{Fig:appendix_cnn}. Note that due to the use of ReLU activation, there is an additional phase in which the dynamics diverges, which is the right portions of Fig.~\ref{Fig:mlp_cnn} with white backgrounds. Both models start to overfit after entering into the chaos phase, indicated by the widening gap between training and test accuracies as well as increasing test loss. 

\begin{figure*}
  \centering
{\includegraphics[width=11cm]{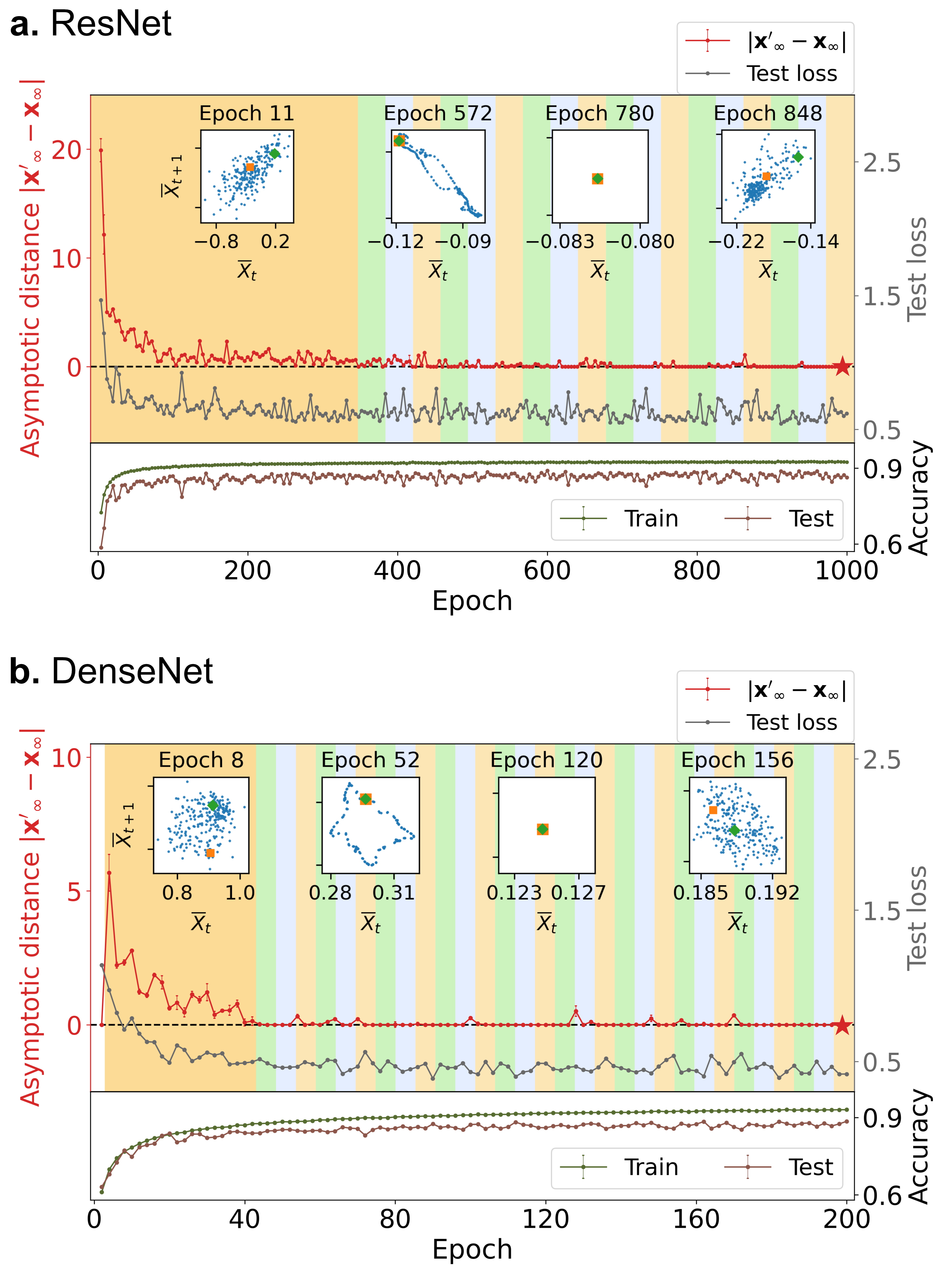}}
  \caption{{\bf Evolution towards edge of chaos during training in modern deep neural networks.} Due to the high computational complexity of Jacobian computation, we use the asymptotic separation $|\pmb x_\infty - \pmb x'_\infty |$ of two nearby trajectories $\pmb x_t$ and $\pmb x'_t$ to directly assess the stability phases numerically instead. The two methods are equivalent in high dimensional systems as discussed and validated in Sec.~\ref{Sec:lya}.
  A. Residual neural network (ResNet)~\cite{he2016deep} and B. Densely connected convolutional networks (DenseNet)~\cite{huang2017densely} are both trained on CIFAR10 dataset. Error bar represents the standard deviation over the test data. The model evolves from the chaotic phase (orange background) towards the edge of chaos where $|\pmb x_\infty - \pmb x'_\infty |$ oscillats around 0 (alternating 3 background colors), while the test accuracies of the model increase. The constant switching among the 3 phases during training also validates our theoretical prediction that the (pseudo)periodic cycle phase (like epoch 572 in A and epoch 52 in B) is extremely narrow in high dimensional systems.
    }
\label{Fig:res_dense}
\end{figure*}

The state-of-art computer vision models use complex  architectures for better accuracy, as well as more refined data augmentation and training algorithms to prevent overfitting. We use the sample implementation of both ResNet~\cite{he2016deep} and the DenseNet~\cite{huang2017densely}, with the modification that adds an upscaling layer after the final convolutional layer, such that this added layer has the same output dimension as the input images (Sec.~\ref{Sec:empirical}). Such modification has no impact on the  accuracies of the models as shown in Table~\ref{Table:res_compare} and ~\ref{Table:dense_compare}. From Fig.~\ref{Fig:res_dense} it is clear that the training process pushes the models towards the edge of chaos as the accuracies improve (more optimal models). Near the edge of chaos, we observe that the model constantly transitions among the three different phases illustrated by the inset Poincare plots, indicating a very narrow phase space the periodic cycle phase as predicted by our theories. More extensive experiments show that the same pattern exists when we vary the depths of the network models (Sec.~\ref{Sec:empirical}).

Note that for very deep neural networks like ResNet and DenseNet, the evaluation of the Jacobian is extremely computationally expensive due to the model sizes. Hence, we use the asymptotic sensitivity of the network operator to initial perturbations as a direct measure of the stability phases. Such quantity is defined as  $|\pmb x_\infty - \pmb x'_\infty|$, where $\pmb x'_0 =\pmb x_0 + \pmb \epsilon$ with $\pmb \epsilon$ being a small value. Numerically this is associated with the maximal Lyapunov exponent that determines the stability/chaos of the operator, with  $|\pmb x_\infty - \pmb x'_\infty|>0$ in the chaotic phase  $|\pmb x_\infty - \pmb x'_\infty|=0$ in the stable phase (both stable periodic cycle phase and single stable fixed point phase).  Since both $|\pmb x_\infty - \pmb x'_\infty|=0$ and  $\frac{1}{\sqrt{N}} \overline{\|\pmb J^*\|}=1$ can be used as the boundary for the edge of chaos, their theoretical relationship can be established through the maximal Lyapunov exponent (see Sec.~\ref{Sec:lya} for theoretical details and empirical validation).

\section{Discussion}
 
 In all of the deep learning models we have tested, their optimal states are near the edge of chaos. In general, the closer it is to the edge of chaos, the higher is the model's performance during the training process, even though they do not always reach the edge. An example is DenseNet with very deep structures, that the model stays in the chaotic phase during training~(Fig.~\ref{Fig:appendix_dense2}B); yet explicit model regularizations like weight decay and dropout used to prevent overfitting can pull the model to the edge of chaos (Fig.~\ref{Fig:appendix_dense2}C), yielding less generalization gap. This indicates that overfitting is associated with chaos. 
 Such relation is also visible from Fig.~\ref{Fig:mlp_cnn}, where the geralization gap between training and testing grows as the models become more chaotic. Hence, our theoretical picture also provides a clear interpretation for two of the most fundamental problems in deep learning: generalization~\cite{jiang2019fantastic} and robustness against adversarial attacks~\cite{goodfellow2014explaining}. Features learnt in the chaotic phase are not stable patterns, rendering the models unable to generalize well and fragile against input perturbations. Extending from this, potentially one can retrieve the `stable' patterns learnt in AI models for explainability studies. 
 
Unlike the popular input-output analysis of neural networks~\cite{novak2018sensitivity,sokolic2017generalization}, here we see that the asymptotic property is the key  to understand the fundamental principles of machine intelligence. Since in high dimensional systems the asymptotic stability is generally intrinsic to the model and independent of the input data, there could be some related fundamental measures without carrying out asymptotic calculations. One example can be found in the special case of single-layer fully connected network. If there is such a measure for generic networks, it can play important roles in studying models that do not permit asymptotic calculations due to their architecture designs or expensive computational costs, especially for the highly complex state-of-art deep learning models.

Our theory is  general with few constraints on the systems of interest, and able to unify  disordered systems and some ordered systems (here the order/disorder refers to the weights of the model, such that a neural network with random weights is disordered while the logistic map with single weight is ordered). Hence, although our experimental validation is carried out only on machine intelligence due to the exact controllability and measurability of those systems, it can be applicable to the biological brain as well. One example is the experimental finding on the biological brain~\cite{beggs2003neuronal},  that the branching factor in neural propagation is close to 1, corresponding to $\frac{1}{N}  \overline {\|\pmb J^*\|}_c^2 \approx 1$. In our theory this is the edge of chaos and  leads to the maximal number of metastable states, which were observed through simulations in~\cite{haldeman2005critical}. With the increasing data and knowledge  in both biological and machine intelligence, more studies can be done in our theoretical framework to extract new principles behind them.

\renewcommand\thefigure{S\arabic{figure}}  
\renewcommand{\thetable}{S\arabic{table}}
\setcounter{figure}{0}   
\setcounter{equation}{0}   
\setcounter{section}{0}   
\renewcommand{\thesection}{S\arabic{section}}
\renewcommand{\theequation}{S\arabic{equation}}

{ \centering \section*{ Appendix}}

\section{Derivation on edge of chaos $\frac{1}{N}\|\pmb J^*\|_c^2 =1$}
\label{Sec:general}

The asymptotic solution of a generic non-linear dynamic operator $\pmb f(\pmb x)$ can be  defined as:
\begin{align}
\pmb x^* = \pmb f(\pmb x^*)
\end{align}
where $\pmb x$ is a vector of dimension $N$. In the case of an image recognition neural network,  $n$ is the number of pixels in the input image.
Hence, it can be also expressed as:
\begin{align}
x_i^* = \pmb f_i(\pmb x^*), i \in \{1,2,3,\cdots, N\}
\end{align}
Assuming each $x_i$ is a stochastic variable representing chaos, we can write:
\begin{align}
x_i^* = \mu_i + \xi_i
\end{align}
where $\mu_i = E[x_i^*]$ is its mean value and $\xi_i$ is the stochastic residual with mean 0 and variance $\sigma_i^2$.

Taking Taylor expansion up to second order around $\pmb x = \pmb\mu$ with $\pmb \mu = [\mu_1,\mu_2,\cdots,\mu_n]$, we have:
\begin{align}
x_i^* &\approx f_i(\pmb x)\bigg\rvert_{\pmb x=\pmb \mu}+\sum_j \frac{\partial f_i(\pmb x)}{\partial x_j}\bigg\rvert_{\pmb x=\pmb \mu} \xi_j + \frac{1}{2}\sum_{j,k} \frac{\partial^2f_i(\pmb x)}{\partial x_j\partial x_k}\bigg\rvert_{\pmb x=\pmb \mu} \xi_j\xi_k\\
&\approx f_i(\pmb \mu)+\sum_j \frac{\partial f_i(\pmb \mu)}{\partial x_j} \xi_j + \frac{1}{2}\sum_{j} \frac{\partial^2f_i(\pmb \mu)}{\partial x_j^2} \xi_j^2
\end{align}
The last step simplifies the notations and assumes the independence/weak correlation of $\xi_j$ and $\xi_k$ for $j\neq k$.

Therefore, the expectation value of $x_i^*$ is
\begin{align}
\mu_i &= E[x_i^*] = f_i(\pmb \mu) +\sum_j \frac{\partial f_i(\pmb \mu)}{\partial x_j} E[\xi_j] + \frac{1}{2}\sum_{j} \frac{\partial^2f_i(\pmb \mu)}{\partial x_j^2} E[\xi_j^2]\\
&=f_i(\pmb \mu) + \frac{1}{2}\sum_{j} \frac{\partial^2f_i(\pmb \mu)}{\partial x_j^2} \sigma_j^2
\end{align}

Similarly we can calculate the variance of $x_i^*$ from the expectation of $x_i^{*2}$ as:
\begin{align}
E[x_i^{*2}] &\approx f_i(\pmb \mu)^2 +\sum_j \left(\frac{\partial f_i(\pmb \mu)}{\partial x_j}\right)^2 \sigma_j^2 +   f_i(\pmb \mu) \sum_{j} \frac{\partial^2f_i(\pmb \mu)}{\partial x_j^2} \sigma_j^2 
\end{align}
which leads to the variance :
\begin{align}
\sigma_i^2 &= E[x_i^{*2}] - (E[x_i^*])^2\\
&=\sum_j \left(\frac{\partial f_i(\pmb \mu)}{\partial x_j}\right)^2 \sigma_j^2 
\end{align}

We then arrive at the average variance over all neurons as:
\begin{align}
\sigma^2&=\frac{1}{N} \sum_i \sigma_i^2= \frac{1}{N}\sum_{i,j} \left(\frac{\partial f_i(\pmb \mu)}{\partial x_j}\right)^2 \sigma_j^2  \\
\end{align}
By the central limit theorem, $\sigma^2$ approximately follows a normal distribution with mean $\frac{1}{N}\sum_{i,j}\left(\frac{\partial f_i(\pmb \mu)}{\partial x_j}\right)^2\cdot \sigma^2$, i.e. a self-consistent equation of itself. Assuming that for large $N$, i.e. in the thermodynamic limit, the value of $\sigma^2$ equals to its expected mean value, we have
\begin{align}
\sigma^2 = \frac{1}{N} \|\pmb J_{\pmb f}^*\|_F^2 \sigma^2
\label{prime}
\end{align}
where $\|\pmb J_{\pmb f}^*\|_F$ is simply the Frobenius norm of the Jacobian for the neural network operator, evaluated at the mean value $\pmb \mu$ of the asymptotic state $\pmb x^*$, i.e.
\begin{align}
\|\pmb J_{\pmb f}^*\|_F^2 = \sum_{i,j} \left(\frac{\partial f_i(\pmb \mu)}{\partial x_j}\right)^2
\label{Eqn:master}
\end{align}

Hence, we can see from Eqn.~\ref{prime} that, the boundary between order and chaos is defined by the asymptotic Jacobian norm equal to 1, i.e.
\begin{align}
\frac{1}{N}\|\pmb J_{\pmb f}^*\|_F^2 = 1
\label{critical2}
\end{align}
For simplicity of notations, we demote $\|\pmb J_{\pmb f}^*\|_F$ as $\|\pmb J^*\|$ and  $\|\pmb J_{\pmb f}\|_F$ as $\|\pmb J\|$ in the paper, and refer $\frac{1}{N}\|\pmb J_{\pmb f}\|_F^2$ as the Jacobian norm. 
In the case of the single layer neural network, we see in the next section that this exact value can be analytically calculated.


%
%

\section{Chaotic boundary in finite dimensions and the maximal Lyapunov exponent}
\label{Sec:lya}

In finite dimensional systems, stable (pseudo)periodic cycles appear between the two phases of stable fixed point and chaos. Then the equation of periodic attractor Eqn.~\ref{Eqn:self_sigma} picks up time dependence:
\begin{align}
\|\pmb \sigma_{l+1}\|^2 = \frac{1}{N}\|\pmb J_l^*\|^2\|\pmb \sigma_{l}\|^2, 
\end{align}
where $l\in[1,2,\cdots,L],$  with $L$ being the period length, and $\|\pmb \sigma_{l+1}\|$ is the amount of chaos at lag $l$ in the period, while $\|\pmb J_l^*\|^2$ is the local Jacobian norm at lag $l$ in the period. Therefore, the average asymptotic behavior of $\|\pmb \sigma\|$ per iteration is related to the geometric average of the asymptotic Jacobian over each $l$:
\begin{align}
\lim_{t\to\infty}  \|\pmb \sigma_{t+1}\|^2 &\approx \frac{1}{N} \left(\prod_{l=1}^L \|\pmb J_l^*\|^2 \right) ^{1/L} \|\pmb \sigma_{t}\|^2 \\
&= \frac{1}{N}  \overline {\|\pmb J^*\|}^2 \|\pmb \sigma_{t}\|^2,
\end{align}
where $\overline {\|\pmb J^*\|} = (\prod_{l=1}^L \|\pmb J_l^*\|)^{1/L}$ is the geometric mean of the asymptotic Jacobian norms. 
One can extend the above to the chaotic phase, by replacing $L$ with a large enough value to approximate the asymptotic behaviors, since $L$ is not well defined in chaos. In fact this is the method in finite time estimation of the maximal Lyapunov exponent.
Therefore, the critical boundary separating periodic cycles and the chaotic phase is then:
\begin{align}
\frac{1}{N}  \overline {\|\pmb J^*\|}_c^2 = 1.
\label{Eqn:p_critical}
\end{align}
When the dimension $N$ is very large, the periodic phase is extremely narrow, such that the one can simply use Eqn.~\ref{critical} to get the analytical result. But for low dimensional systems with a broad periodic phase such as the logistic map, Eqn.~\ref{Eqn:p_critical} is needed, together with the exact asymptotic values of $\pmb x_t$ in the periodic cycle or chaotic attractor.

Since the LHS of Eqn.~\ref{Eqn:p_critical} defines the average multiplicative factor of $\|\pmb \sigma\|^2$, it directly translates into the maximal Lyapunov exponent $\gamma$ as:
\begin{align}
\gamma &= \lim_{\tau \to\infty} \frac{1}{\tau}\sum_{t=0}^{\tau-1}\ln \frac{|\delta {\pmb x}_{t+1}|}{|\delta {\pmb x}_{t}|}\\
& = \lim_{\tau \to\infty} \frac{1}{2\tau}\sum_{t=0}^{\tau-1}\ln \frac{\delta \|\pmb \sigma_{t+1}\|^2}{\delta \|\pmb \sigma_{t}\|^2} \\
&=\ln (\frac{1}{\sqrt{N}}\overline{\|\pmb J^*\|})
\label{Eqn:gamma}
\end{align}

Commonly finite time estimation of the maximal Lyapunov exponent can also be calculated by linearizing the system followed by calculating its maximal eigenvalue $\lambda_1$ of the linearized matrix: $\gamma = \frac{1}{\tau}\ln |\lambda_1|$, with $\lambda_1$ being the largest absolute eigenvalue of the matrix:
\begin{align}
\underline{\pmb J} = \prod_{t=T+1}^{T+\tau}\pmb J_t,
\label{Eqn:original_lya}
\end{align}
where both $T$ and $\tau$ are large to ensure convergence. In this case, $|\lambda_1|$ is simply the spectral radius $\rho_{T,\tau}$ of this matrix. 
For large dimension $N$, the spectral radius $\rho$ of an Jacobian matrix is approximately the same as its normalized Jacobian norm as discussed perviously.  Assuming the local Jacobian $\pmb J_t$ at different time $t$ are random and independent of each other, and $N$ is large, we  have:
\begin{align}
& \rho_t = \frac{1}{\sqrt{N}} \|\pmb J_t\|\\
\Longrightarrow \ &\lambda_1=\rho_{T,\tau} =  \frac{1}{\sqrt{N}} \left\| \prod_{t=T+1}^{T+\tau}\pmb J_t \right\|  =  N^{-\frac{\tau}{2}}  \prod_{t=T+1}^{T+\tau}\left\|\pmb J_t \right\| \\
\Longrightarrow \ &\gamma =  \frac{1}{\tau}\ln \|\lambda_1\| =\ln (\frac{1}{\sqrt{N}}\overline{\|\pmb J^*\|}),
\end{align}
which is exactly Eqn.~\ref{Eqn:gamma}. 

\begin{figure*}
  \centering
{\includegraphics[width=17cm]{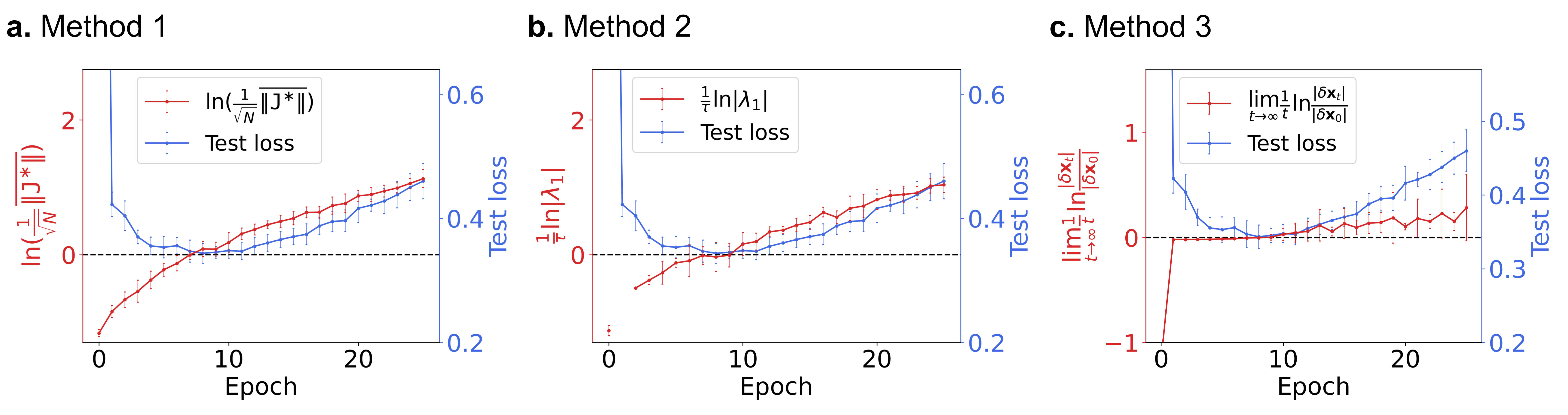}}
  \caption{{\bf Equivalence of different methods for maximal Lyapunov exponent in MLP.} The three methods have similar patterns during the transition to chaos. Method 2 is unable to yield a result due to the numerical precision in computing the eigenvalue of the multiplicative matrix. The error bars represent the standard deviation over 10 different experimental runs.}
\label{Fig:appendix_three_methods_mlp}
\end{figure*}

\begin{figure*}
  \centering
{\includegraphics[width=17cm]{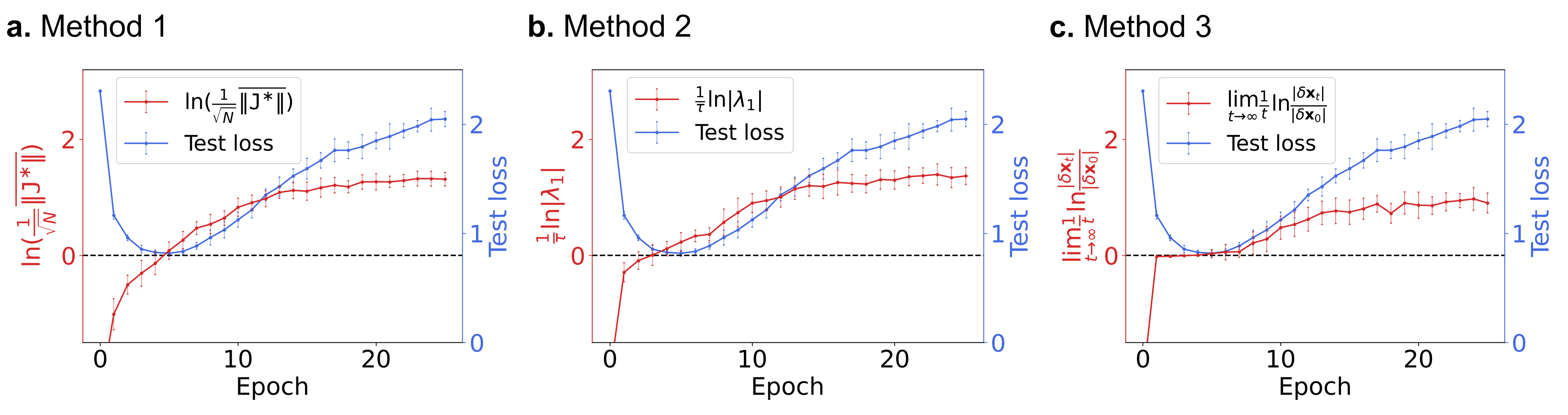}}
  \caption{{\bf Equivalence of different methods for maximal Lyapunov exponent for CNN.} Results are similar to MLP.}
\label{Fig:appendix_three_methods_cnn}
\end{figure*}

This means our method of estimating $\gamma$ from the Jacobian norm's geometric mean in Eqn.~\ref{Eqn:gamma} (method 1) is equivalent to finding the maximal absolute eigenvalue of the multiplicative linearization in Eqn.~\ref{Eqn:original_lya} (method 2). Figure.~\ref{Fig:appendix_three_methods_mlp}AB and Figure.~\ref{Fig:appendix_three_methods_cnn}AB demonstrate this equivalence for both MLP and CNN. For our method using Eqn.~\ref{Eqn:gamma}, we assume ergodicity of the system, i.e. a single ergodic attractor for almost any initial state. Therefore the sample average is equivalent to the time average, and we use sample average with each sample at the final iteration, such that the convergence to the attractor is better. For the second method using Eqn.~\ref{Eqn:original_lya}, we use time average due to the definition of this quantity. Specifically, we discard the first $T/2$ iterations of each input image to ensure convergence, and use the rest of the iterations to estimate a $\gamma$ value for each input. In practice, the second method tend to have issues on numerical accuracies using large $\tau$ values. This is because the neural networks trained on empirical datasets leads to  $\pmb J_t$ matrix are not independent and i.i.d, such that the multiplication of a large number of them in Eqn.~\ref{Eqn:original_lya} leads to a matrix with many zero elements due to numerical precision, which in turn result in inaccurate $\lambda_1$ values. 

For very deep neural networks, the Jacobian matrix is extremely computationally intensive to compute. Hence, for most of those networks in our experiments, we use a third method of directly calculating the trajectory separations between two close initial inputs. Simply, this method 3 calculates the value $|\delta \pmb x_{T} | = | \pmb x_{T} - \pmb x'_T|$, which we denote as `{\it asymptotic distance  $|\pmb x'_\infty-\pmb x_\infty|$}'. This quantity is also related to the numerical estimation of the maximal Lyapunov exponent:
\begin{align}
\gamma \approx \frac{1}{T} \ln \frac{|\delta \pmb x_{T} |}{|\delta \pmb x_{0} |},
 \end{align}
where $\delta \pmb x_{0} = | \pmb x_{0} - \pmb x'_0|$ with $\pmb x'_0 = \pmb x_0 + \pmb\epsilon$. This relation is exact when $\pmb \epsilon \to \pmb 0$.
Here we chose $\pmb\epsilon$ to be a very small value. The results are shown in  Figure.~\ref{Fig:appendix_three_methods_mlp}C and Figure.~\ref{Fig:appendix_three_methods_cnn}C. The technical details of the three methods are as follows:
\begin{itemize}
\item{Number of images used in method 1 and 3 are 100.  Method 2 uses 1 image. }
\item{Iteration number $T$  is 500 for method 1 and 3. For method 2 we  use $T=200$, and only use the second half of the iterations to compute the results. This is because when $T$ is too large, one can easily encounter `no convergence problem' in our eigenvalue calculations.}
\item{Method 1 and 2 both stop simulation early when the length of the output is smaller than $1^{-10}$ or larger than $1^{10}$.}
\end{itemize}

Note that in all of our experiments, we validate the phases using Poincare maps to double check the stability phases indicated by the theoretical predictions.

\section{Logistic map and mutual information}
\label{Sec:log_map}

In Fig.~\ref{Fig:phase_diagram}C, the dynamic operator is constructed 20 layers of logistic maps, i.e. 
 \begin{align}
x_{t+1} = \underbrace{g_r \circ \cdots  \circ g_r}_\text{20 times}(x_{t}),
\end{align} 
where $g_r(x) = rx(1-x)$.
To get the attractor at different $r$ values, we first iterate the dynamics by 1000 times, and then use the next 1000 iterations to plot the attractors in Fig.~\ref{Fig:phase_diagram}C.

\begin{figure*}
  \centering
{\includegraphics[width=16cm]{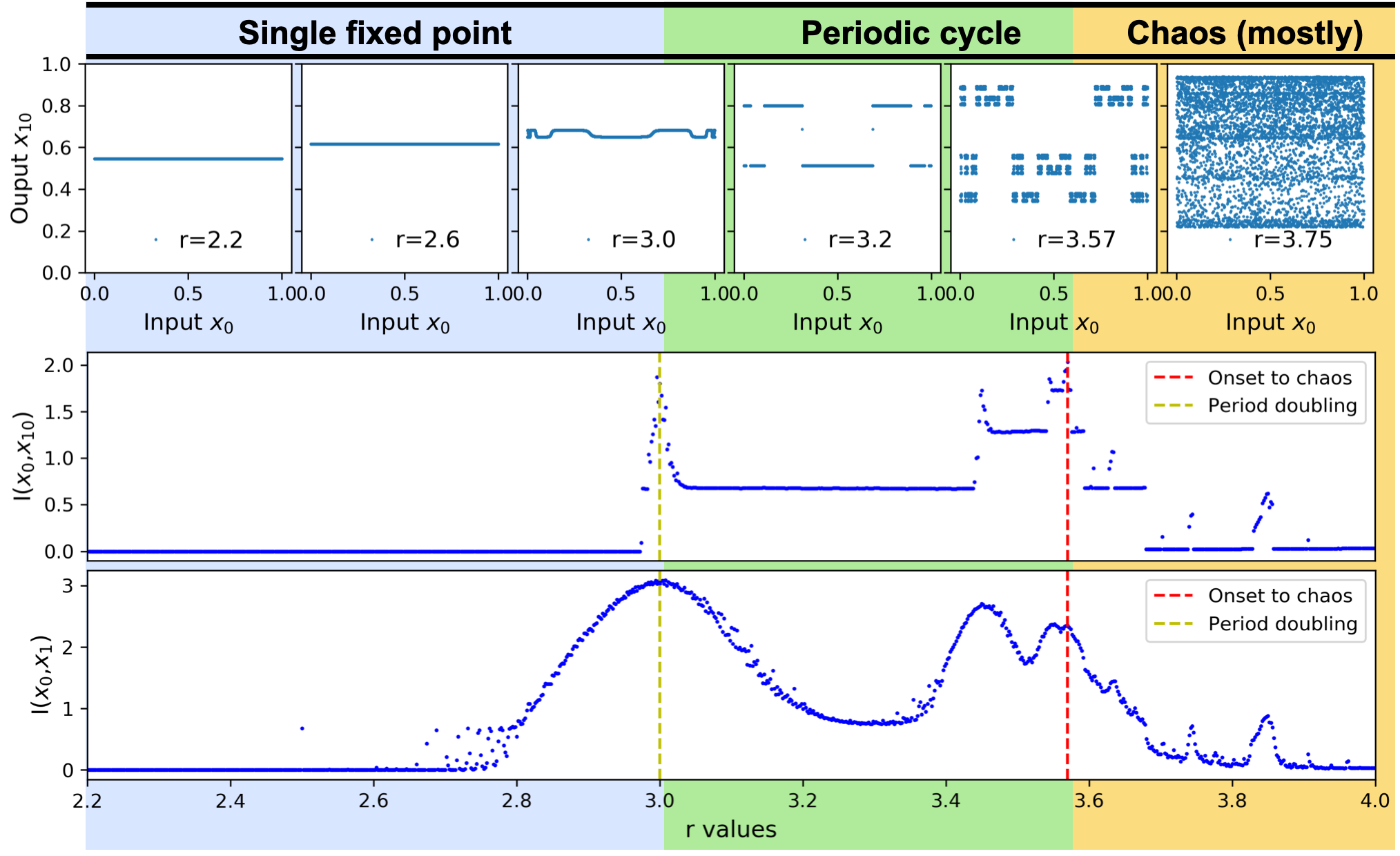}}
  \caption{{\bf Mutual information between input $x_0$ and asymptotic output $x_{10}$ of the 20-layer logistic map toy operator.}  The top figure shows the scatter plots of input and outputs for different $r$ values.
  The centre figure shows the mutual information $I(x_0,x_{10})$ calculated from the top figures at different $r$ values transitioning across the three phases, and the bottom figure shows mutual information $I(x_0,x_{1})$.
}
\label{Fig:Log_maps}
\end{figure*}

To investigate the information processing by this toy logistic map operator, we examine the mutual information between the input value $x_0$ and its asymptotic output $x_\infty$. Usually at the 10th iteration of the dynamics is already converged to the attractor. Hence, we only need to measure the mutual information between the input $x_0$ and the output $x_{10}$ at the 10th iteration. It is well-known that as the $r$ value increases from 0, the logistic map goes through a transition from single fixed point to period doubling cascades, until it reaches the first onset to chaos at $r\approx 3.57$. This is reflected clearly in the input/output diagram in Fig.~\ref{Fig:Log_maps} which demonstrates the three phases.

The mutual information (MI) measure $I(x_0,x_{10})$ between $x_0$ and $x_{10}$ in the logistic map can be decomposed into:
\begin{align}
I(x_0,x_{10}) &= H(x_{10}) - H(x_{10}|x_0),
\end{align}
where $H(x_{10})$ is then entropy of $x_{10}$ and $H(x_{10}|x_0)$ is the conditional entropy of $x_{10}$ given $x_0$.
For deterministic functions that maps $x_0$ to $x_{10}$, the second term on the right is theoretically 0. But in chaotic phase this can only be true with perfect accuracy on the value of $x_0$, which is impossible in practice. Therefore, we numerically measure $I(x_0,x_{10})$ by putting them into 500 bins of size $0.002$, which is similar to a measurement accuracy of $0.002$. Note that both $x_0$  and $x_{10}$ are confined in the range (0, 1). To ensure accurate statistics, we sample 4,000,000 pairs of ($x_0,x_{10}$) uniformly in the range $0<x_0<1$, such that each bin on average has 16 samples.

Because the mapping from input to output is deterministic, $H(x_{10}|x_0)=0$ in the order phase. It is known that in the order phase before reaching the first onset to chaos at $r\approx 3.57$, the logistic map dynamics converges to a stable cycle of period $L$ for any input $x$. With 10 iterations, the system is reaching the asymptotic behavior of $L$-period cycle, such that $x_{10}$ converges towards the $L$ different periodic values. In other words, $x_{10}$ values are concentrated in only $L$ different discrete states, as seen in Fig.~\ref{Fig:Log_maps} for $r=3.2$ and $r=3.57$. Therefore the effective mutual information is equivalent to the entropy from those $L$ states, i.e. $I(x_0,x_{10})=H(x_{10})\approx \ln L$ if we use the $0^{th}$ order Renyi entropy for analytical simplicity. This implies that the information processing capability of $f_r(x)$ is maximal when $L$ is maximal. Since $L$ is maximal at the end of period doubling, i.e. onset to chaos at $r\approx 3.57$, we expect the mutual information peaks at this point, which is indeed the case shown in Fig.~\ref{Fig:Log_maps}.

However, in the chaotic phase, the assumption that $H(x_{10}|x_0)=0$ fails in practice, because an infinitesimally small   change in the input $x_0$ will result in huge change in $x_{10}$. In this case $H(x_{10}|x_0)>0$, so that $I(x_0,x_{10})<H(x_{10})$, and the information processing capability of the network generally decreases from the onset of chaos $r\approx3.57$. The high mutual information between the input $x_0$ and asymptotic states $x_\infty$ also infers high mutual information between $x_0$ and $x_1$, as the two mutual information measures are highly correlated as shown in Fig.~\ref{Fig:Log_maps}.

A side finding is that both $I(x_0,x_{1})$ and $I(x_0,x_{10})$ are locally maximal at the point of period doubling, $r=3$ as shown in Fig.~\ref{Fig:Log_maps} for instance. This is because the convergence to the attractor is extremely slow near this critical phase transition point, resulting in large number of different $x_t$ values for finite $t$. 
However, for high dimensional systems, such transition to different periodic length is hard to find since the whole periodic cycle phase is already extremely narrow in the phase diagram as shown in Fig.~\ref{Fig:phase_diagram}B. Hence, for neural networks one can roughly consider the edge of chaos is the whole periodic cycle phase.


\section{Experimental validation on various deep neural networks}
\label{Sec:empirical}

\subsection{Classification accuracy for different networks}

\begin{figure*}[h!]
  \centering
{\includegraphics[width=12cm]{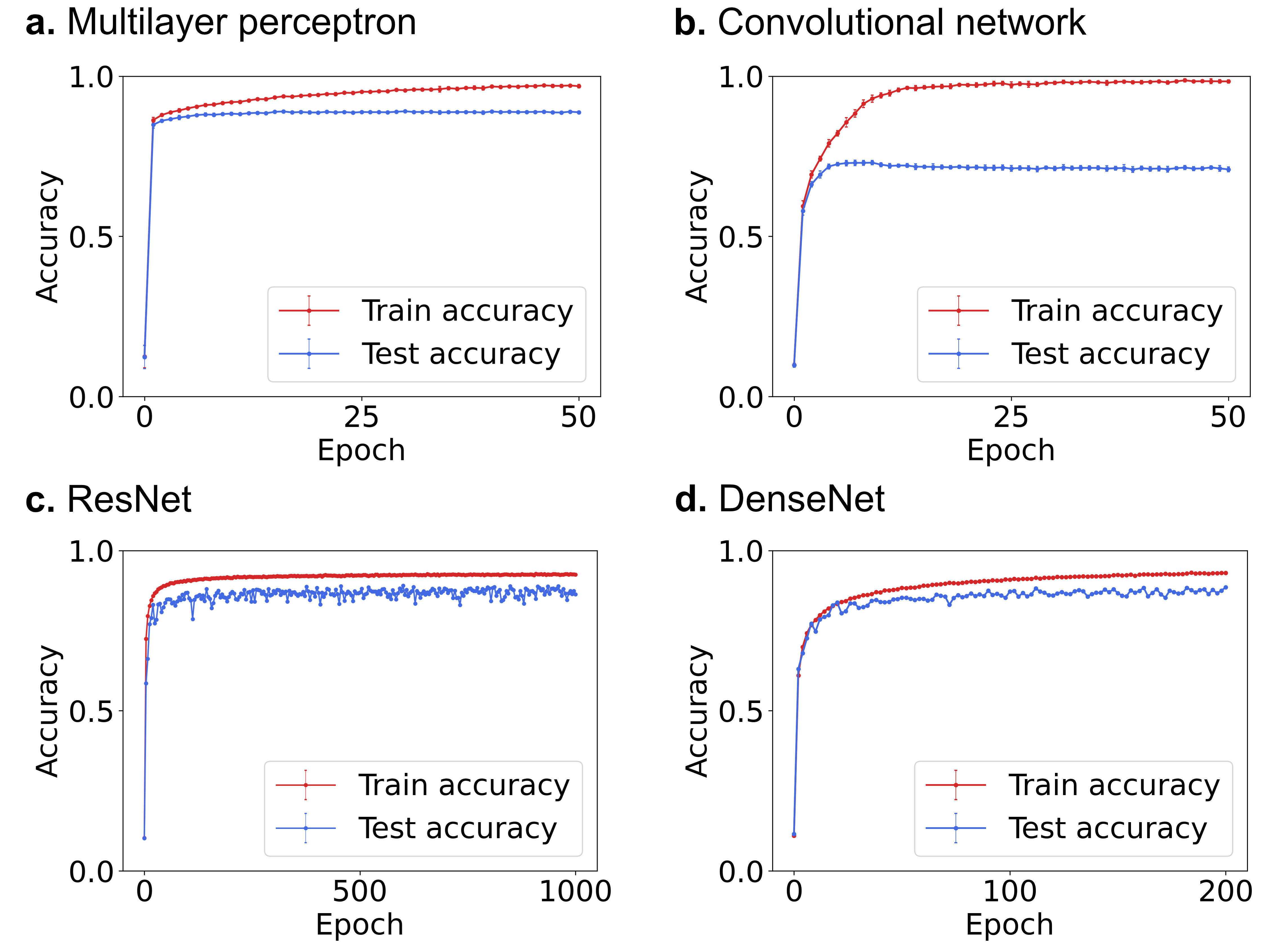}}
  \caption{{\bf Classification accuracy for different networks.}}
\label{Fig:appendix_classification_accuracy}
\end{figure*}

Fig.~\ref{Fig:appendix_classification_accuracy} shows the classification accuracy for networks depicted in Fig.~\ref{Fig:mlp_cnn} and Fig.~\ref{Fig:res_dense}. Here in Fig.~\ref{Fig:appendix_classification_accuracy} longer training epochs are shown until  both training and test accuracies are saturated.

\subsection{Jacobian norm's geometric mean for ResNet and DenseNet}

\begin{figure*}
  \centering
  {\includegraphics[width=15cm]{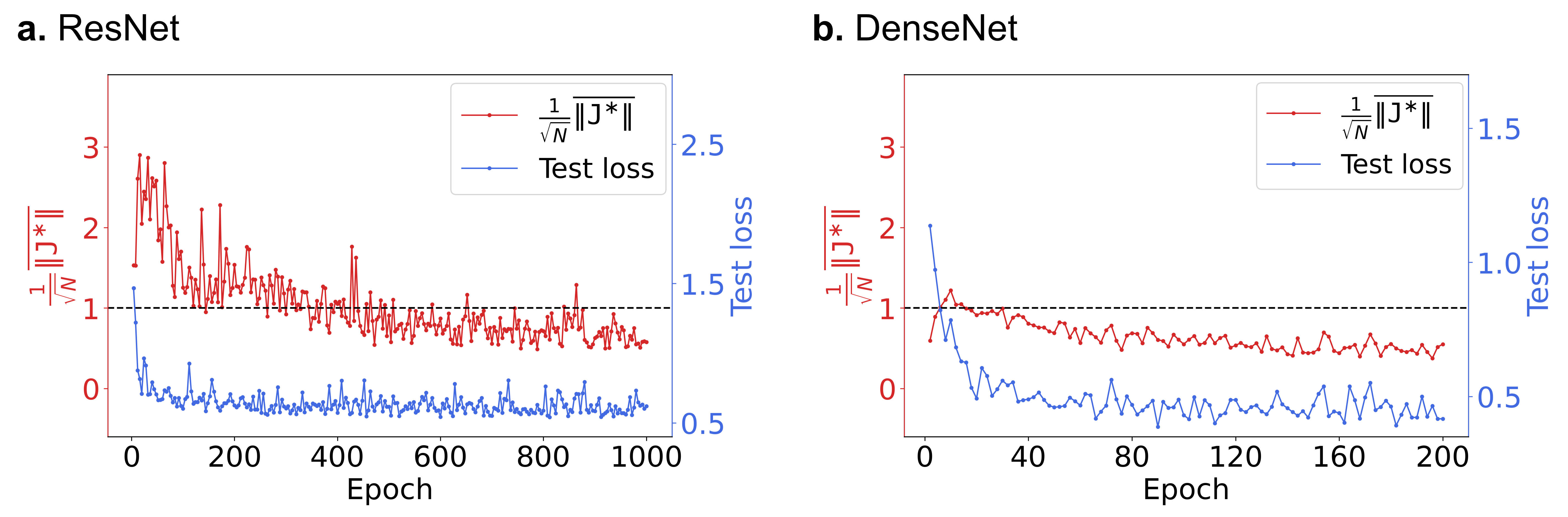}}
  \caption{{\bf Jacobian norm's geometric mean for ResNet in Fig.~\ref{Fig:res_dense}A and DenseNet in Fig.~\ref{Fig:res_dense}B.}}
\label{Fig:appendix_jacobian_norm}
\end{figure*}

Fig.~\ref{Fig:appendix_jacobian_norm} shows the Jacobian norm's geometric mean for networks in Fig.~\ref{Fig:res_dense}. Both ResNet and DenseNet plots in Fig.~\ref{Fig:appendix_jacobian_norm}  indicate  evolutions towards more stability, as validated by the Poincare plots in Fig.~\ref{Fig:res_dense}. For ResNet, the Jacobian norm $\frac{1}{\sqrt{N}} \overline{\|\pmb J\|}$ oscillates around 1 in later epochs, indicating edge of chaos. For DenseNet, the model is mostly in the slightly stable phase, close to the results from Poincare maps with slightly deviations. Such deviation could be due to the correlations in the weight matrices, such that the i.i.d.  assumption in our theory does not hold strictly. But the 3-phase phenomenon is not affected by such correlations.

\subsection{Different deep neural network layers and structures}

We carry out experiments on various network architectures and training techniques with Tensorflow 2.1.0. In the training of the networks, we always use ‘Adam’ optimizer~\cite{kingma2014adam} as it is the most commonly used optimizer in computer vision tasks. Learning rate is the default setting in Tensorflow. Activation function used throughout is rectified linear unit (ReLU) except for the output layer which is softmax. All of the multilayer perceptrons (MLPs) are trained on Fashion MNIST dataset as it is simple enough for MLPs to achieve good accuracy. The other models are trained on CIFAR10 \footnote{\url{https://www.cs.toronto.edu/~kriz/cifar.html}} dataset which is a standard dataset for sophisticated computer vision models.

\begin{figure*}[h!]
  \centering
{\includegraphics[width=18cm]{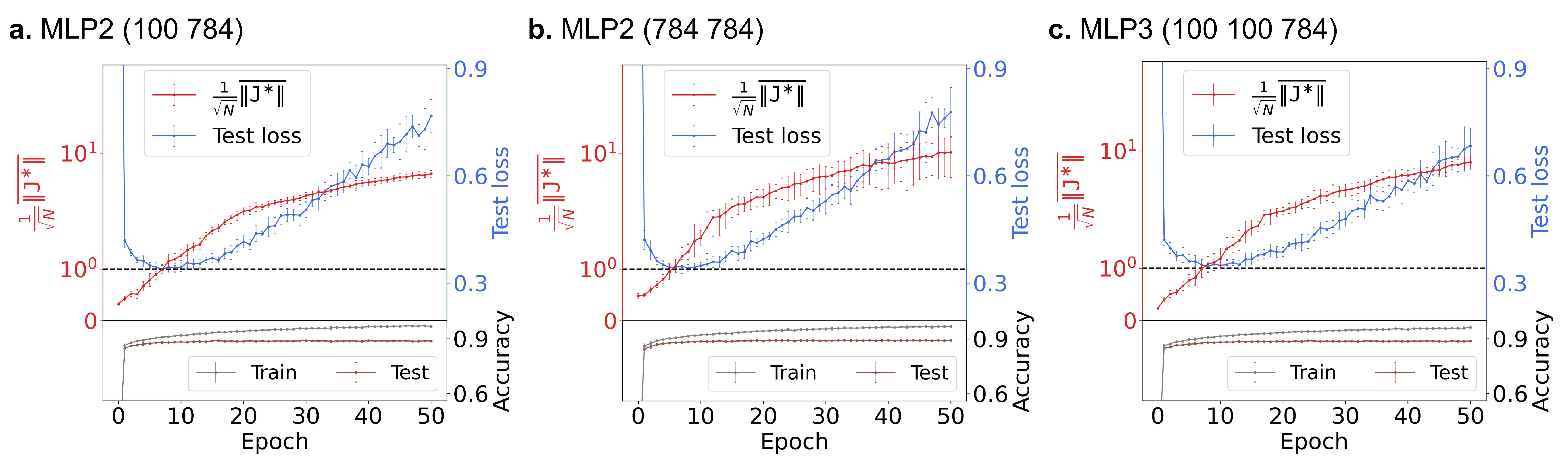}}
  \caption{{\bf Jacobian norm's geometric mean for various MLPs.} Error bars represent standard error over 10 repeat experiments.}
\label{Fig:appendix_mlp}
\end{figure*}

The MLP in Fig.~\ref{Fig:mlp_cnn}A has two hidden layers with 100 and 784 nodes each. We also experimented with different versions of MLP that have different number of hidden layers and number of nodes in each layer. Note that the final hidden layer is fixed at 784 nodes - same dimension as the input layer. The models in Fig.~\ref{Fig:appendix_mlp} have structure details in their titles. For instance, MLP2 (100 784) refers to two hidden layers with 100 and 784 nodes each. Table.~\ref{table:appendix_mlp} shows the test accuracy for the models in Fig.~\ref{Fig:appendix_mlp}.

\begin{table*}[h!]
  \caption{Test accuracy for various MLPs.}
  \label{table:appendix_mlp}
  \centering
  \begin{tabular}{l|c|c|c}
    \hline
                  & MLP2 (100 784) & MLP2 (784 784)  & MLP3 (100 100 784)\\
    \hline
    Test accuracy & 89.0\%         & 89.4\%          & 88.9\%  \\
    \hline
  \end{tabular}
\end{table*}

\begin{figure*}[h!]
  \centering
{\includegraphics[width=18cm]{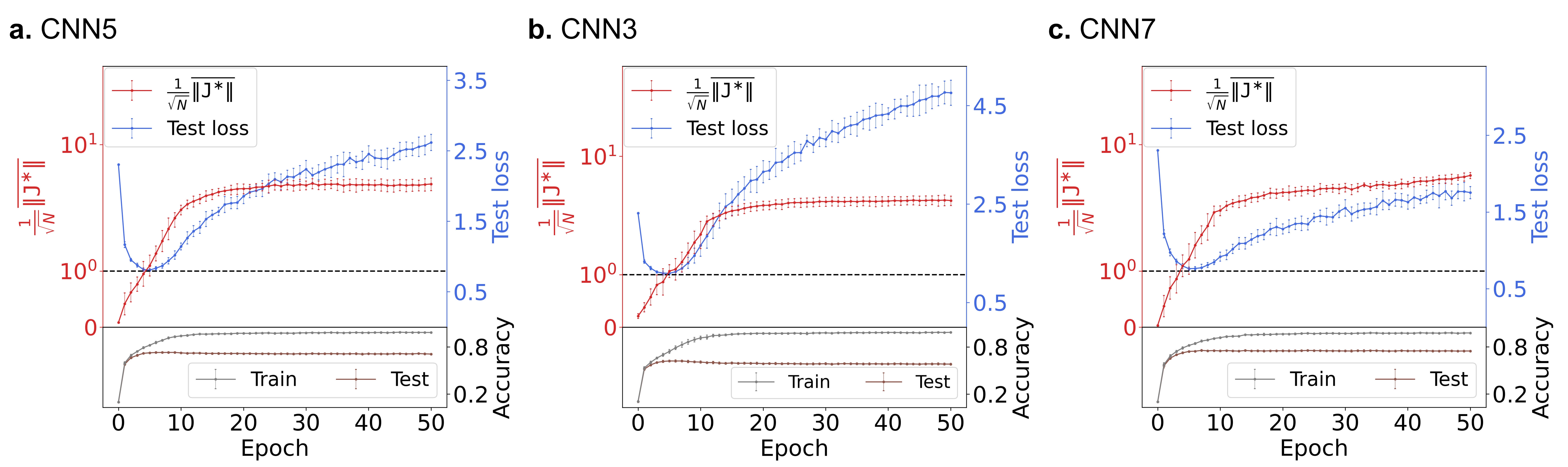}}
  \caption{{\bf Jacobian norm's geometric mean for various CNNs.}}
\label{Fig:appendix_cnn}
\end{figure*}

We modify the convolutional neural network (CNN) implementation from Keras \footnote{\url{https://github.com/keras-team/keras/blob/master/examples/cifar10_cnn.py}}. To keep the input-output dimension of the extracted dynamical operator the same, we upscale the output from the final convolutional layer by adding a composite function of three consecutive operations: DepthToSpace \footnote{\url{https://www.tensorflow.org/api_docs/python/tf/nn/depth_to_space}}, followed by a $3\times3$ convolution with 3 filters and a ReLU activation. DepthToSpace is used to make the channel dimension of input and output the same, and $3\times3$ convolution with 3 filters is used to make the number of channels of input and output the same. This kind of composite function is added to ResNet and DenseNet as well for the same reason. 
With the composite function, the network in Fig.~\ref{Fig:mlp_cnn}B has 5 convolutional layers and we refer it as CNN5 in Fig.~\ref{Fig:appendix_cnn}. Table.~\ref{table:appendix_cnn} shows the test accuracy for the models in Fig.~\ref{Fig:appendix_cnn}. To see the influence on model performance with the modification we made, we also compare our results to the standard CNN implementation in Keras, denoted as CNN5*\ . The test accuracy of CNN5 and CNN5*\ show that the model performance is little affected by the composite function.

\begin{table*}[h]
  \caption{Test accuracy for various CNNs.}
  \label{table:appendix_cnn}
  \centering
  \begin{tabular}{ l | c | c | c | c}
    \hline
                  & CNN5    & CNN3      & CNN7    & CNN5*\      \\
    \hline
    Test accuracy & 73.0\%  & 62.2\%    & 75.5\%  & 74.7\%      \\
    \hline
  \end{tabular}
  \label{Table:cnn_compare}
\end{table*}

\begin{figure*}[h!]
  \centering
{\includegraphics[width=18cm]{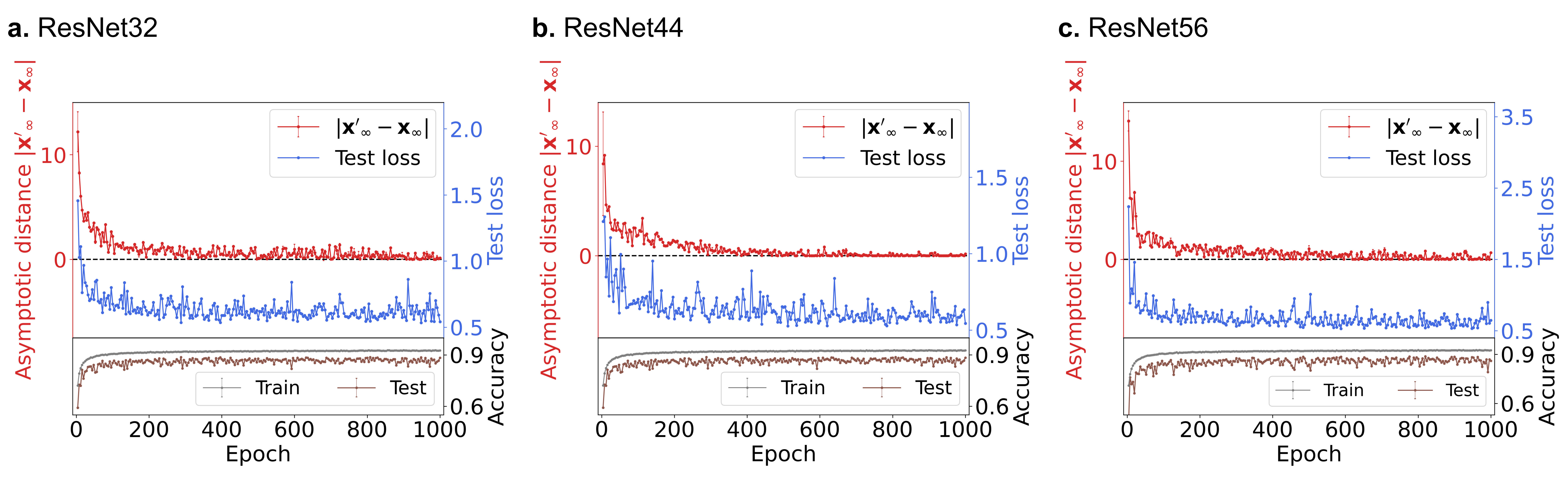}}
  \caption{{\bf Stability evolution for various ResNets.}}
\label{Fig:appendix_res}
\end{figure*}

For ResNet, in addition to adding the composite upscaling function to the network implementation from Keras \footnote{\url{https://github.com/keras-team/keras/blob/master/examples/cifar10_resnet.py}}, we also remove the global pooling layer, since our final convolutional layer has only 3 channels which are too few for global pooling to work. ResNet in Fig.~\ref{Fig:res_dense}A has 20 layers(ResNet20), and Fig.~\ref{Fig:appendix_res} shows the results for ResNet with 32, 44 and 56 layers. Table.~\ref{table:appendix_res} shows the test accuracy for the models in Fig.~\ref{Fig:appendix_res} and Fig.~\ref{Fig:res_dense}A. We also compare our results to the standard Keras implementation of ResNet, denoted as ResNet20*. The test accuracy of ResNet20 and ResNet20*\ are again very close.

\begin{table*}[h]
  \caption{Test accuracy for ResNets.}
  \label{table:appendix_res}
  \centering
  \begin{tabular}{l|c|c|c|c|c}
    \hline
                  & ResNet32  & ResNet44  & ResNet56  & ResNet20  & ResNet20*\ \\
    \hline
    Test accuracy & 88.7\%    & 88.9\%    & 88.8\%    & 89.0\%      & 88.9\%     \\
    \hline
  \end{tabular}
  \label{Table:res_compare}
\end{table*}

\begin{figure*}[h!]
  \centering
{\includegraphics[width=18cm]{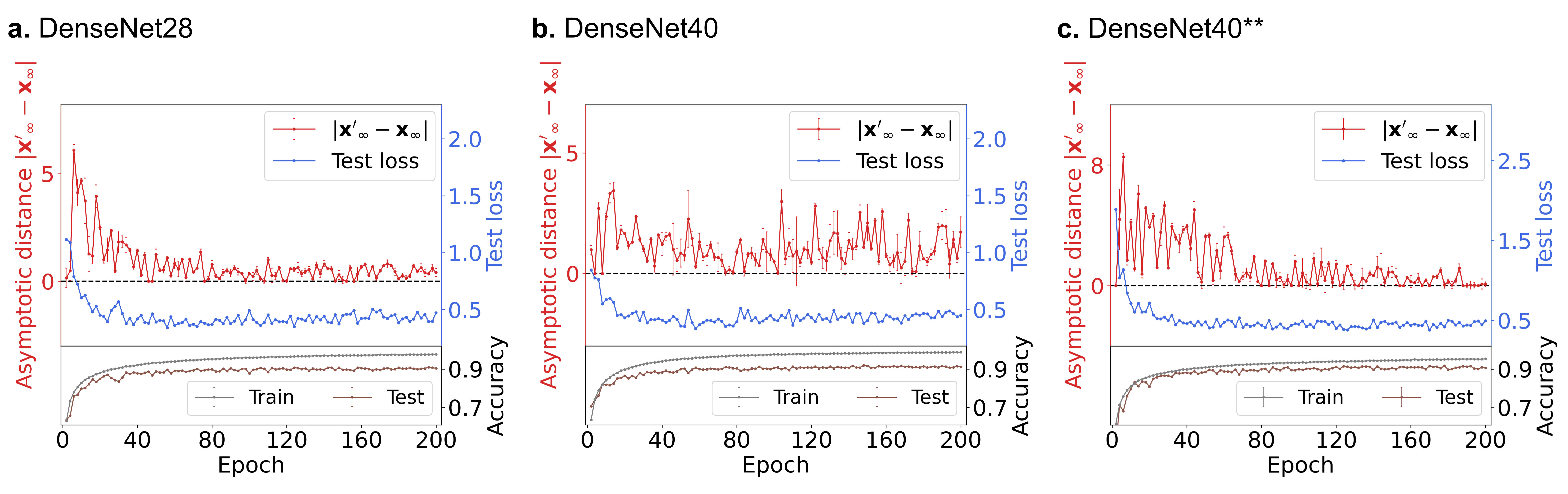}}
  \caption{{\bf Stability evolution for various DenseNets.}}
\label{Fig:appendix_dense2}
\end{figure*}

For DenseNet, we adopt the implementation from Microsoft \footnote{\url{https://github.com/microsoft/samples-for-ai/blob/master/examples/keras/DenseNet/densenet.py}}. The final global pooling layer is also removed for the same reason as ResNet. The DenseNet in Fig.~\ref{Fig:res_dense} has 16 layers (DenseNet16), and Fig.~\ref{Fig:appendix_dense2} shows the results for DenseNet with 28 and 40 layers. Table.~\ref{table:appendix_dense2} shows the test accuracy for the models in Fig.~\ref{Fig:appendix_dense2} and Fig.~\ref{Fig:res_dense}B. DenseNet16*\ is again the sample implementation. The test accuracy of DenseNet16 and DenseNet16*\ show our modification has little impact on the model performance.
 For DenseNet with 40 layers, it remains in the chaotic phase even after long training time as seen in Fig.~\ref{Fig:appendix_dense2}B. To investigate the effect of model regularization,  we add a weight decay of $10^{-4}$ and dropout layers with dropout rate 0.2 to this model, and label it as DenseNet40** as shown in Fig.~\ref{Fig:appendix_dense2}C. Compared to Fig.~\ref{Fig:appendix_dense2}B, it can be seen that regularization brings the model to the edge of chaos from the chaotic phase, while yielding smaller generalization gap between training and test accuracies.

\begin{table*}[h]
  \caption{Test accuracy for DenseNets.}
  \label{table:appendix_dense2}
  \centering
  \begin{tabular}{l|c|c|c|c|c}
    \hline
                  & DenseNet28  & DenseNet40  & DenseNet40**\ & DenseNet16   & DenseNet16*\ \\
    \hline
    Test accuracy & 91.3\%      & 92.0\%       & 91.9\%      & 88.5\%       & 88.4\%\\
    \hline
  \end{tabular}
  \label{Table:dense_compare}
\end{table*}

\bibliography{draft}

\begin{thebibliography}{10}

\bibitem{munoz2018colloquium}
Miguel~A Munoz.
\newblock Colloquium: Criticality and dynamical scaling in living systems.
\newblock {\em Reviews of Modern Physics}, 90(3):031001, 2018.

\bibitem{Beggs11167}
John~M. Beggs and Dietmar Plenz.
\newblock Neuronal avalanches in neocortical circuits.
\newblock {\em Journal of Neuroscience}, 23(35):11167--11177, 2003.

\bibitem{Fraiman2009}
Daniel Fraiman, Pablo Balenzuela, Jennifer Foss, and Dante~R. Chialvo.
\newblock Ising-like dynamics in large-scale functional brain networks.
\newblock {\em Phys. Rev. E}, 79:061922, Jun 2009.

\bibitem{bak59self}
P~Bak, C~Tang, and K~Wiesenfeld.
\newblock Self-organized criticality: an explanation of 1/f noise.
\newblock {\em Phys. Rev. Lett}, 59:381, 1987.

\bibitem{Crutchfield2008}
David~P. Feldman, Carl~S. McTague, and James~P. Crutchfield.
\newblock The organization of intrinsic computation: Complexity-entropy
  diagrams and the diversity of natural information processing.
\newblock {\em Chaos: An Interdisciplinary Journal of Nonlinear Science},
  18(4):043106, 2008.

\bibitem{HUBERMAN1986376}
B.A. Huberman and T.~Hogg.
\newblock Complexity and adaptation.
\newblock {\em Physica D: Nonlinear Phenomena}, 22(1):376 -- 384, 1986.
\newblock Proceedings of the Fifth Annual International Conference.

\bibitem{beggs2003neuronal}
John~M Beggs and Dietmar Plenz.
\newblock Neuronal avalanches in neocortical circuits.
\newblock {\em Journal of neuroscience}, 23(35):11167--11177, 2003.

\bibitem{haldeman2005critical}
Clayton Haldeman and John~M Beggs.
\newblock Critical branching captures activity in living neural networks and
  maximizes the number of metastable states.
\newblock {\em Physical review letters}, 94(5):058101, 2005.

\bibitem{bertschinger2004real}
Nils Bertschinger and Thomas Natschl{\"a}ger.
\newblock Real-time computation at the edge of chaos in recurrent neural
  networks.
\newblock {\em Neural computation}, 16(7):1413--1436, 2004.

\bibitem{LEGENSTEIN2007323}
Robert Legenstein and Wolfgang Maass.
\newblock Edge of chaos and prediction of computational performance for neural
  circuit models.
\newblock {\em Neural Networks}, 20(3):323 -- 334, 2007.
\newblock Echo State Networks and Liquid State Machines.

\bibitem{sherrington1975solvable}
David Sherrington and Scott Kirkpatrick.
\newblock Solvable model of a spin-glass.
\newblock {\em Physical review letters}, 35(26):1792, 1975.

\bibitem{nishimori2001statistical}
Hidetoshi Nishimori.
\newblock {\em Statistical physics of spin glasses and information processing:
  an introduction}.
\newblock Number 111. Clarendon Press, 2001.

\bibitem{sompolinsky1988chaos}
Haim Sompolinsky, Andrea Crisanti, and Hans-Jurgen Sommers.
\newblock Chaos in random neural networks.
\newblock {\em Physical review letters}, 61(3):259, 1988.

\bibitem{Toyoizumi:2011aa}
T.~Toyoizumi.
\newblock Beyond the edge of chaos: Amplification and temporal integration by
  recurrent networks in the chaotic regime.
\newblock {\em Physical Review E}, 84(5), 2011.

\bibitem{almeida1978stability}
J~R~L de~{Almeida} and D~J {Thouless}.
\newblock Stability of the sherrington-kirkpatrick solution of a spin glass
  model.
\newblock {\em Journal of Physics A}, 11(5):983--990, 1978.

\bibitem{tao2010random}
Terence Tao, Van Vu, Manjunath Krishnapur, et~al.
\newblock Random matrices: Universality of esds and the circular law.
\newblock {\em The Annals of Probability}, 38(5):2023--2065, 2010.

\bibitem{EDELMAN1997203}
Alan Edelman.
\newblock The probability that a random real gaussian matrix haskreal
  eigenvalues, related distributions, and the circular law.
\newblock {\em Journal of Multivariate Analysis}, 60(2):203 -- 232, 1997.

\bibitem{GREBOGI632}
CELSO GREBOGI, EDWARD OTT, and JAMES~A. YORKE.
\newblock Chaos, strange attractors, and fractal basin boundaries in nonlinear
  dynamics.
\newblock {\em Science}, 238(4827):632--638, 1987.

\bibitem{feigenbaum1976universality}
MJ~Feigenbaum.
\newblock Universality in complex discrete dynamics.
\newblock Technical report, LA-6816-PR, LASL Theoretical Division Annual Report
  July 1975---September, 1976.

\bibitem{Schuecker:2018aa}
Jannis Schuecker.
\newblock Optimal sequence memory in driven random networks.
\newblock {\em Physical Review X}, 8(4), 2018.

\bibitem{kingma2014adam}
Diederik~P Kingma and Jimmy Ba.
\newblock Adam: A method for stochastic optimization.
\newblock {\em arXiv preprint arXiv:1412.6980}, 2014.

\bibitem{xiao2017fashionmnist}
Han Xiao, Kashif Rasul, and Roland Vollgraf.
\newblock Fashion-mnist: a novel image dataset for benchmarking machine
  learning algorithms, 2017.

\bibitem{krizhevsky2009learning}
Alex Krizhevsky, Geoffrey Hinton, et~al.
\newblock Learning multiple layers of features from tiny images.
\newblock 2009.

\bibitem{he2016deep}
Kaiming He, Xiangyu Zhang, Shaoqing Ren, and Jian Sun.
\newblock Deep residual learning for image recognition.
\newblock In {\em Proceedings of the IEEE conference on computer vision and
  pattern recognition}, pages 770--778, 2016.

\bibitem{huang2017densely}
Gao Huang, Zhuang Liu, Laurens Van Der~Maaten, and Kilian~Q Weinberger.
\newblock Densely connected convolutional networks.
\newblock In {\em Proceedings of the IEEE conference on computer vision and
  pattern recognition}, pages 4700--4708, 2017.

\bibitem{jiang2019fantastic}
Yiding Jiang, Behnam Neyshabur, Hossein Mobahi, Dilip Krishnan, and Samy
  Bengio.
\newblock Fantastic generalization measures and where to find them.
\newblock {\em arXiv preprint arXiv:1912.02178}, 2019.

\bibitem{goodfellow2014explaining}
Ian~J Goodfellow, Jonathon Shlens, and Christian Szegedy.
\newblock Explaining and harnessing adversarial examples.
\newblock {\em arXiv preprint arXiv:1412.6572}, 2014.

\bibitem{novak2018sensitivity}
Roman Novak, Yasaman Bahri, Daniel~A Abolafia, Jeffrey Pennington, and Jascha
  Sohl-Dickstein.
\newblock Sensitivity and generalization in neural networks: an empirical
  study.
\newblock {\em arXiv preprint arXiv:1802.08760}, 2018.

\bibitem{sokolic2017generalization}
Jure Sokolic, Raja Giryes, Guillermo Sapiro, and Miguel Rodrigues.
\newblock Generalization error of invariant classifiers.
\newblock In {\em Artificial Intelligence and Statistics}, pages 1094--1103.
  PMLR, 2017.

\end{thebibliography}

\bibliographystyle{unsrt}

\end{document}